# Privacy-Preserved Big Data Analysis Based on Asymmetric Imputation Kernels and Multiside Similarities

Bo-Wei Chen

School of Information Technology, Monash University, AU

**Abstract**

This study presents an efficient approach for incomplete data classification, where the entries of samples are missing or masked due to privacy preservation. To deal with these incomplete data, a new kernel function with asymmetric intrinsic mappings is proposed in this study. Such a new kernel uses three-side similarities for kernel matrix formation. The similarity between a testing instance and a training sample relies not only on their distance but also on the relation between the testing sample and the centroid of the class, where the training sample belongs. This reduces biased estimation compared with typical methods when only one training sample is used for kernel matrix formation. Furthermore, centroid generation does not involve any clustering algorithms. The proposed kernel is capable of performing data imputation by using class-dependent averages. This enhances Fisher Discriminant Ratios and data discriminability. Experiments on two open databases were carried out for evaluating the proposed method. The result indicated that the accuracy of the proposed method was higher than that of the baseline. These findings thereby demonstrated the effectiveness of the proposed idea.

*Keywords: Incomplete data analysis, partial similarity, multiside similarity, privacy preservation, kernel ridge regression (KRR), missing values, data imputation, kernel method, cloud computing, data analytics*

## 1. Introduction

Incomplete data analysis has become a research hotspot with the recent increasing demand for big data processing, in addition to complexity problems due to huge volumes. Take Internet of Things for example. Data collected by large-scale sensor networks could reach trillions in the future. However, when sensors fail, defective data are recorded in the dataset, subsequently resulting in biased estimation. In cloud computing, the same problem arises not merely because of erroneous samples, but because of privacy protection. Sensitive personal data, such as health records, faces, and names, are intentionally removed from the original data to avoid being maliciously manipulated [1-3]. These defective or masked data subsequently form an incomplete dataset. A systematic approach for conquering incomplete data problems is evitable.

Actually, as far back as the earlier 1970s [4], this problem already aroused much attention from scientists when they dealt with nonresponses in surveys. Missing data usually result from physically mechanical failure or human factors. For instance, subjects neglect questions in questionnaires deliberately or unintentionally.

To handle such a problem, data imputation is one of the most commonly used techniques when scientists face missing data. Imputation is a statistical term that describes the process of replacing missing data with substituted values [5]. Data deletion is a straightforward and easy approach for incomplete data processing. Once a missing value appears in an attribute/field/variable, the entire attribute is removed from the data. However, if the deleted attribute contains key information, discriminability may degrade. To improve such a problem, single imputation and multiple imputation were developed. The former replaces a sample that contains missing entries with a similar instance. This type of approaches include hot decks (i.e., inserted values are selected based on the same dataset), cold decks (i.e., insertion is derived from another dataset), stochastic regression (e.g., interpolation), subspace-based reconstruction [6, 7], and fixed-value replacement (e.g., means or medians). Single imputation has been widely used because of its efficiency and simplicity although it generates values that cannot fully reflect uncertainty. Multiple imputation accommodates such a point by considering the insertion as a set of stochastic numbers, or values with random noise. Thus, inserted values along with unmissing entries satisfy the distribution of complete data. Multiple imputation is performed by randomly taking values from a set of stochastic numbers and then filling in all the missing fields in the dataset $L$ times [8]. Therefore, $L$ complete datasets are duplicated based on the original incomplete dataset. Each complete dataset is individually analyzed. The final result comes from the fusion of $L$ analyses since there are $L$ datasets. Multiple imputation was proposed by Donald B. Rubin, and this technique created a milestone for incomplete data analysis.

Compared with single imputation, multiple imputation yields results that approximate actual data distributions. Nevertheless, to approach real distributions, Monte Carlo methods [9] or Markov Chain Monte Carlo (MCMC) were frequently adopted to simulate theoretical scenarios by using finite sampling. For small datasets, multiple imputation is an effective and efficient solution to incomplete data analysis. When the volume of data becomes excessively large, generation of $L$ complete datasets requires a delicate mechanism.

Much research related to machine learning was examined for the performance comparison of data imputation. For example, Troyanskaya *et al.* [6] examined data imputation by using $K$-Nearest Neighbors (KNNs) and Singular Value Decomposition (SVD). The result showed that when the percentage of missing values reached 20%, the normalized root-mean-squared errors (RMSEs) were below 0.18, both lower than those methods with zero-padding and average-filling techniques. Farhangfar *et al.* [10] compared several imputation methods, including hot decks, polytomous regression, Naïve Bayesian regressors, and mean insertion. Classification on incomplete data was carried out by testing C4.5, Support Vector Machines (SVMs), Naïve Bayesian classifiers, and KNNs. SVMs with polynomial kernels and mean insertion outperformed SVMs with Radial Basis Functions (RBFs) and the other classifiers. In big incomplete data analytics, both effectiveness and efficiency should receive highly attention since big data imputation jointly considers accuracy and complexity. Anagnostopoulos and Triantafillou [11] investigated the scalability of imputation. They evaluated accuracy and efficiency of big data imputation by analyzing whether or not a group of clustered computers worked better than a single powerful machine. They devised weighted KNNs and sequential multivariate regression for imputation. The experiment indicated their proposed distributed method, i.e., weighted imputation, was better than naïve MapReduce mechanisms and superior to centralized processing. Kung and Wu [12] focused on kernelized methods and devised a zero-padding method for single imputation. After filling in missing entries with substituted values, the Partial Fisher Discriminant-Ratio of each attribute was calculated and sorted, so that discriminant attributes could be selected.

Regarding big data analysis, two types of machine-learning approaches are frequently investigated in literature. One is divide-and-conquer [13-16], and the other is incremental [17-23]. The former focuses on recursively dividing the original problem into subproblems. The solution to each subproblem is computed in parallel and then combined to give a solution for the original problem [24]. To relieve computational burdens and enhance the performance, entire data are usually divided into smaller sets so that individual computers can share the load. Unlike the divide-and-conquer methods, the latter type does not involve processing subproblems in a distributed way, but the data are sequentially fed into the same classifier. Incremental learning requires a mechanism for updating model parameters without retraining the entire model. It is also called online learning. Both divide-and-conquer and incremental methods can be applied to big incomplete data analysis.

Unlike the abovementioned strategies, this study presents a novel mechanism for big incomplete data analysis, where Fast Kernel Ridge Regression (Fast KRR) powered by the proposed asymmetric imputation kernel is used during data processing. Fast KRR aims at rapid data processing, and asymmetric imputation kernels focus on imputation, where privacy data can be masked and filled in with substituted values by using the proposed asymmetric imputation kernel.

The advantage of the proposed methodology is that class information is considered during imputation. This is conducive to discriminability enhancement when the system fills in missing entries with substituted values. Second, kernel matrices no longer rely on the similarity between pairwise samples. Instead, three-side similarities are adopted, where the third side comes from the centroid of a class. This improves robustness against incomplete information. Furthermore, centroid generation does not involve any clustering algorithms.

The rest of this paper is organized as follows. Section II introduces the proposed asymmetric data imputation kernel. Sections III then describes the details of Fast KRR. Next, Section IV summarizes the performance of the proposed method and the analytic results. Conclusions are finally drawn in Section V.

## 2. Asymmetric Imputation Kernel

This study followed the method “Kernel Approach to Incomplete Data Analysis (KAIDA)” [12] with several enhanced modifications for data imputation. The original imputation mechanism [12] is fulfilled by zero-padding a vector that contains missing values. Let **B** represent a mask that performs imputation. Then,

$$\mathbf{B}_{\mathbf{x}}(\iota)=\begin{cases}1 & \text{if } \mathbf{x}_{\iota} \text{ is given}\\ 0 & \text{otherwise}\end{cases} \tag{1}$$

and

$$\tilde{\mathbf{x}}=\mathbf{x}\otimes\mathbf{B}_{\mathbf{x}} \tag{2}$$

where $\iota$ denotes the $\iota$–th dimension of a vector, and $\otimes$ is the element-wise operator, i.e., Hadamard operators.

However, zero-padding results in a decrease of Partial Fisher Discriminant-Ratios, shown as follows. Let $F$ denote the Partial Fisher Discriminant-Ratio. Thus,

$$F_{\iota}=\frac{\left(\mu_{\iota}^{+}-\mu_{\iota}^{-}\right)^{2}}{\left(\sigma_{\iota}^{+}\right)^{2}+\left(\sigma_{\iota}^{-}\right)^{2}} \tag{3}$$

where $\mu$ and $\sigma$ are respectively the partial mean and the partial variance of the ι–th dimension. Classes “+” and “-” respectively refer to positive and negative samples. The Fisher Discriminant-Ratio measures the discrepancy between two distributions. When the ratio becomes higher, the discrepancy is larger. This implies higher discriminability.

Assume that the number of unmissing entries in the ι–th dimension of class “+” is $n_{\iota}^{+}$ . The incremental computation of the mean and the variance is

$$\mu_\iota^+\left[n_\iota^+ + 1\right] = \mu_\iota^+\left[n_\iota^+\right] + \frac{\eta - \mu_\iota^+\left[n_\iota^+\right]}{n_\iota^+ + 1} \tag{4}$$

and

$$\left(\sigma_\iota^+\left[n_\iota^+ + 1\right]\right)^2 = n_\iota^+\left(\left(n_\iota^+ - 1\right)\left(\sigma_\iota^+\left[n_\iota^+\right]\right)^2 + n_\iota^+\left(\mu_\iota^+\left[n_\iota^+\right] - \mu_\iota^+\left[n_\iota^+ - 1\right]\right)^2 + \left(\eta - \mu_\iota^+\left[n_\iota^+ + 1\right]\right)^2\right) \tag{5}$$

where $[\cdot]$ is the iteration, and $\eta$ is the value of an entry or an attribute in an instance.

If $\eta$ is zero as (1) does, (4) and (5) yield

$$\mu_\iota^+\left[n_\iota^+ + 1\right] = \mu_\iota^+\left[n_\iota^+\right] - \frac{\mu_\iota^+\left[n_\iota^+\right]}{n_\iota^+ + 1} < \mu_\iota^+\left[n_\iota^+\right]$$

and

$$\left(\sigma_\iota^+\left[n_\iota^+ + 1\right]\right)^2 = \left(n_\iota^+\right)^{-1}\left(\left(n_\iota^+ - 1\right)\left(\sigma_\iota^+\left[n_\iota^+\right]\right)^2 + n_\iota^+\left(\mu_\iota^+\left[n_\iota^+\right] - \mu_\iota^+\left[n_\iota^+ - 1\right]\right)^2 + \left(\mu_\iota^+\left[n_\iota^+ + 1\right]\right)^2\right) > \left(\sigma_\iota^+\left[n_\iota^+\right]\right)^2.$$

This causes a decrease of (3). Namely,

$$F_\iota\left[n+1\right] < F_\iota\left[n\right]. \tag{6}$$

This indicates that when more zero-padding performs, discriminability in the ι–th dimension diminishes.

As our focus is upon data classification, discriminability is the first priority when imputed data are generated. To prevent Partial Fisher Discriminant-Ratios from decreasing, let $\eta = \mu_\iota^+\left[n_\iota^+\right]$. Subsequently, (4) and (5) respectively become

$$\mu_\iota^+\left[n_\iota^+ + 1\right] = \mu_\iota^+\left[n_\iota^+\right]$$

and

$$\left(\sigma_\iota^+\left[n_\iota^+ + 1\right]\right)^2 = \frac{n_\iota^+ - 1}{n_\iota^+}\left(\sigma_\iota^+\left[n_\iota^+\right]\right)^2.$$

This does not decrease Partial Fisher Discriminant-Ratios since the nominator remains unchanged, and the denominator is smaller. In terms of Signal-to-Noise Ratios (SNRs) [25], the denominator, i.e., variance, is noise. When variance diminishes, SNRs are enhanced.

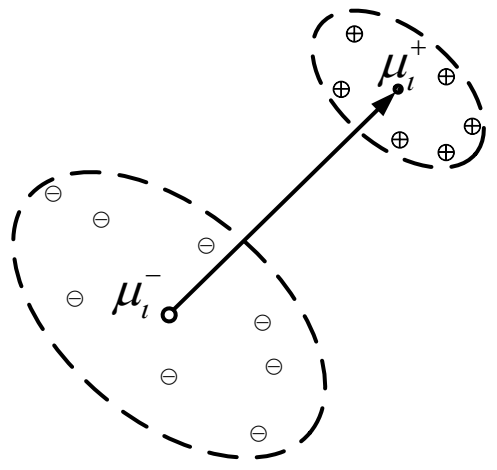

Fig. 1. Illustration of Fisher Discriminant-Ratios

Typically, a kernel matrix measures the similarity between two vectors. However, when the system, e.g., [12], calculates such a matrix, no class information is used. Take the cosine-similarity function for example. The kernel matrix is formed by calculating

$$K_{\text{Cosine}}\left(\mathbf{x}_i,\mathbf{x}_j\right)=\frac{\mathbf{x}_i^{\mathrm{T}}\mathbf{x}_j}{\left\|\mathbf{x}_i\right\|\left\|\mathbf{x}_j\right\|} \tag{7}$$

where $i$ and $j$ respectively specify the indices of two instances, T means the transpose operator, and $\|\cdot\|^2$ calculates the $\mathcal{L}_2$-norm distance. When these two vectors contain missing entries, it involves nonvectorial similarities and results in biased estimation. Thus, the Masked Partial-Cosine (MPC) function mentioned in [12] was used for nonvectorial similarities.

$$K_{\text{MPC}}\left(\tilde{\mathbf{x}}_i,\tilde{\mathbf{x}}_j\right)=\frac{\tilde{\mathbf{x}}_i^{\mathrm{T}}\tilde{\mathbf{x}}_j}{\left\|\tilde{\mathbf{x}}_i\right\|\left\|\tilde{\mathbf{x}}_j\right\|} \tag{8}$$

$$\begin{cases}\tilde{\mathbf{x}}_i=\mathbf{x}_i\otimes\mathbf{B}_{\mathbf{x}_i}\otimes\mathbf{B}_{\mathbf{x}_i}\\ \tilde{\mathbf{x}}_j=\mathbf{x}_j\otimes\mathbf{B}_{\mathbf{x}_i}\otimes\mathbf{B}_{\mathbf{x}_i}\end{cases}. \tag{9}$$

The idea of the cosine similarity in (8) can be further extended to Pearson’s correlations, where the similarity is standardized.

$$K_{\text{MPP}}\left(\tilde{\mathbf{x}}_i,\tilde{\mathbf{x}}_j\right)=\frac{\left(\tilde{\mathbf{x}}_i-\tilde{\mathbf{z}}_r\right)^{\mathrm{T}}\left(\tilde{\mathbf{x}}_j-\tilde{\mathbf{z}}_q\right)}{\left\|\tilde{\mathbf{x}}_i-\tilde{\mathbf{z}}_r\right\|\left\|\tilde{\mathbf{x}}_j-\tilde{\mathbf{z}}_q\right\|} \tag{10}$$

where $\tilde{\mathbf{z}}_r$ and $\tilde{\mathbf{z}}_q$ respectively represent the estimated centroids of training class $r$ and $q$. Besides, $\tilde{\mathbf{x}}_i$ and $\tilde{\mathbf{x}}_j$ are the instances of class $r$ and $q$, respectively. Equation (10) works well during the training phase because class information is given in the dataset. Nonetheless, the testing phase does not provide class information for testing inputs. This requires estimation of centroids, like $K$-means or spectral clustering. It could create much computational time when data sizes are huge.

To alleviate such a problem, since class information is conducive to discriminability, this study proposes a new mechanism, where class information is indirectly embedded in the kernel matrix. Rather than using two-side similarities, this study adopts a three-side comparison, where the third one is the centroid of a training class. Such a third party $\tilde{\mathbf{z}}$ can provide an additional clue and compensates the bias when missing entries appear in the two-side similarity, applicable both at the training and testing stages. Regardless of the class information of $\tilde{\mathbf{x}}_i$, the system can use the training information $\tilde{\mathbf{x}}_j$ to obtain $\tilde{\mathbf{z}}_q$. In other words, either at the training stage or the testing stage, $\tilde{\mathbf{x}}_i$ needs to consider both $\tilde{\mathbf{x}}_j$ and $\tilde{\mathbf{z}}_q$. For efficiency, centroids are computed simply based on the mean of the training samples in this work, where missing attributes are directly ignored.

The following equation shows the details of the proposed Masked Partial Three-Side Cosine (MPT) function.

$$K_{\mathrm{MPT}}\left(\tilde{\mathbf{x}}_i,\tilde{\mathbf{x}}_j\right)=\frac{\tilde{\mathbf{x}}_i^{\mathrm{T}}\left(\tilde{\mathbf{x}}_j+\tilde{\mathbf{z}}_q\right)}{\left\|\tilde{\mathbf{x}}_i\right\|\left\|\tilde{\mathbf{x}}_j+\tilde{\mathbf{z}}_q\right\|}. \tag{11}$$

At the testing stage, $\tilde{\mathbf{x}}_i$ signifies a testing vector, and $\tilde{\mathbf{x}}_j$ represents a training sample, of which the class information is present. As these centroids are derived from the training data with missing entries, therefore,

$$\mu_\iota^q\left[n_\iota^q+1\right]=\begin{cases}\mu_\iota^q\left[n_\iota^q\right] & \text{if } \iota \text{ is missing}\\ \mu_\iota^q\left[n_\iota^q\right]+\dfrac{\eta_\iota^q-\mu_\iota^q\left[n_\iota^q\right]}{n_\iota^q+1} & \text{otherwise}\end{cases} \tag{12}$$

where $\mu^q$ refers to $\tilde{\mathbf{z}}_q$ in (11). This creates imputation with a class-dependent average in the same attribute. The effect of $\tilde{\mathbf{x}}_j+\tilde{\mathbf{z}}_q$ indicates that the similarity between $\tilde{\mathbf{x}}_i$ and $\tilde{\mathbf{x}}_j$ should also consider the similarity between $\tilde{\mathbf{x}}_i$ and the class centroid of $\tilde{\mathbf{x}}_j$. Second, the missing entries are filled with $\tilde{\mathbf{z}}_q$.

Equation (11) can be extended into polynomial kernels and RBFs, i.e.,

$$K_{\mathrm{MPT}}\left(\tilde{\mathbf{x}}_i,\tilde{\mathbf{x}}_j\right)=\left(1+\tau^{-2}\frac{\tilde{\mathbf{x}}_i^{\mathrm{T}}\left(\tilde{\mathbf{x}}_j+\tilde{\mathbf{z}}_q\right)}{\left\|\tilde{\mathbf{x}}_i\right\|\left\|\tilde{\mathbf{x}}_j+\tilde{\mathbf{z}}_q\right\|}\right)^p \tag{13}$$

and

$$K_{\mathrm{MPT}}\left(\tilde{\mathbf{x}}_i,\tilde{\mathbf{x}}_j\right)=\exp\left(-\left(2\tau^2\right)^{-1}\left\|\frac{\tilde{\mathbf{x}}_i}{\left\|\tilde{\mathbf{x}}_i\right\|}-\frac{\tilde{\mathbf{x}}_j+\tilde{\mathbf{z}}_q}{\left\|\tilde{\mathbf{x}}_j+\tilde{\mathbf{z}}_q\right\|}\right\|^2\right) \tag{14}$$

where $\tau^2$ is the variance of the distribution, and $p$ is the kernel order.

## 3. Kernel Ridge Regression

KRR extends linear regression techniques, in which a ridge parameter is imposed on the objective function to regularize a model [25]. KRR has two types of operation modes. One is intrinsic space, and the other is empirical space. After a kernel function $\phi$ maps features onto hyperspace, intrinsic-space computation yields favorable complexity if the number of data $N$ is far larger than the feature dimension $M$. Otherwise, empirical-space operations should be used.

### 3.1. Intrinsic Space

Intrinsic space is used to describe dispersion matrices, also called intrinsic covariance matrices, computed based on intrinsic dimensions of a sample [25, 26]. Let $\{(\mathbf{x}_i, y_i) | i = 1, \ldots, N\}$ denote a pair of an $M$-dimensional feature vector $\mathbf{x}_i$ and its corresponding label $y_i$, where $i$ specifies the indices of $N$ training samples. The objective of a linear regressor is to minimize the following cost function of least squares errors (LSEs).

$$\min_{\mathbf{u},b} E_{\text{KRR}}(\mathbf{u},b) = \min_{\mathbf{u},b} \left\{ \sum_{i=1}^{N} \left( \mathbf{u}^{\text{T}} \phi(\mathbf{x}_i) + b - y_i \right)^2 + \rho \|\mathbf{u}\|^2 \right\} \tag{15}$$

where $E_{\text{KRR}}$ is the cost function, $\mathbf{u}$ represents a $J$-by-1 weight vector, $\phi(\mathbf{x}_i)$ denotes the intrinsic-space $J$-by-1 feature vector of $\mathbf{x}_i$, $b$ is a bias term, and $\rho$ specifies the ridge parameter. Notably, $J$ is the degree of intrinsic space when feature vectors are transformed by a kernel function.

Equation (15) can be rewritten as a matrix form, i.e.,

$$E_{\text{KRR}}(\mathbf{u},b) = \left\| \mathbf{\Phi}^{\text{T}} \mathbf{u} + b\mathbf{e} - \mathbf{y} \right\|^2 + \rho \|\mathbf{u}\|^2. \tag{16}$$

Individually differentiating (16) with respect to $\mathbf{u}$ and $b$ followed by zeroing both equations gives

$$\mathbf{u} = \left[ \mathbf{\Phi}\mathbf{\Phi}^{\text{T}} + \rho \mathbf{I} \right]^{-1} \mathbf{\Phi} \left[ \mathbf{y} - b\mathbf{e} \right] \tag{17}$$

and

$$b = \frac{1}{N} \left( \mathbf{e}^{\text{T}} \mathbf{y} - \mathbf{e}^{\text{T}} \mathbf{\Phi}^{\text{T}} \mathbf{u} \right). \tag{18}$$

Notice that $\mathbf{K} = \mathbf{\Phi}^{\text{T}}\mathbf{\Phi}$ instead of $\mathbf{\Phi}\mathbf{\Phi}^{\text{T}}$ mentioned in (17). The solution to (17) and (18) can be obtained by solving a system of linear equations in (19).

$$\begin{bmatrix} \mathbf{u} \\ b \end{bmatrix} = \begin{bmatrix} \mathbf{\Phi}\mathbf{\Phi}^{\text{T}} + \rho \mathbf{I} & \mathbf{\Phi}\mathbf{e} \\ \mathbf{e}^{\text{T}} \mathbf{\Phi}^{\text{T}} & N \end{bmatrix}^{-1} \begin{bmatrix} \mathbf{\Phi}\mathbf{y} \\ \mathbf{e}^{\text{T}} \mathbf{y} \end{bmatrix}. \tag{19}$$

*3.2. Empirical Space*

Empirical space refers to dispersion matrices, also called empirical covariance matrices, computed based on the number of samples [25, 27]. According to the Learning Subspace Property in [25], the weight vector **u** has the following relation with **Φ** and an unknown $N$-dimensional vector **a**.

$$\mathbf{u} = \mathbf{\Phi a}\,. \tag{20}$$

Combining (16) and (20) yields

$$E'_{\mathrm{KRR}}\left(\mathbf{a}, b\right) = \left\|\mathbf{Ka} + b\mathbf{e} - \mathbf{y}\right\|^2 + \rho\mathbf{a}^{\mathrm{T}}\mathbf{Ka}. \tag{21}$$

Rearranging the equations after differentiating (21) with respect to **a** and $b$ yields

$$\mathbf{a} = \left[\mathbf{K} + \rho\mathbf{I}\right]^{-1}\left[\mathbf{y} - b\mathbf{e}\right] \tag{22}$$

and

$$b = \frac{\mathbf{y}^{\mathrm{T}}\left[\mathbf{K} + \rho\mathbf{I}\right]^{-1}\mathbf{e}}{\mathbf{e}^{\mathrm{T}}\left[\mathbf{K} + \rho\mathbf{I}\right]^{-1}\mathbf{e}}\;. \tag{23}$$

## 4. Experimental Result

Experiments on two datasets were carried out for evaluating the performance. The information of these datasets is listed in Table I. The first column shows the name. The rest columns specify the number of classes, samples, and dimensions, respectively. Dataset “MIT/BIH ECG” is available at PhysioNet (www.physionet.org), and “ALL-AML” is downloaded from the UC Irvine (UCI) Machine Learning Repository (archive.ics.uci.edu/ml/). The datasets show two typical data, where both $N > M$ and $M > N$ are presented, respectively.

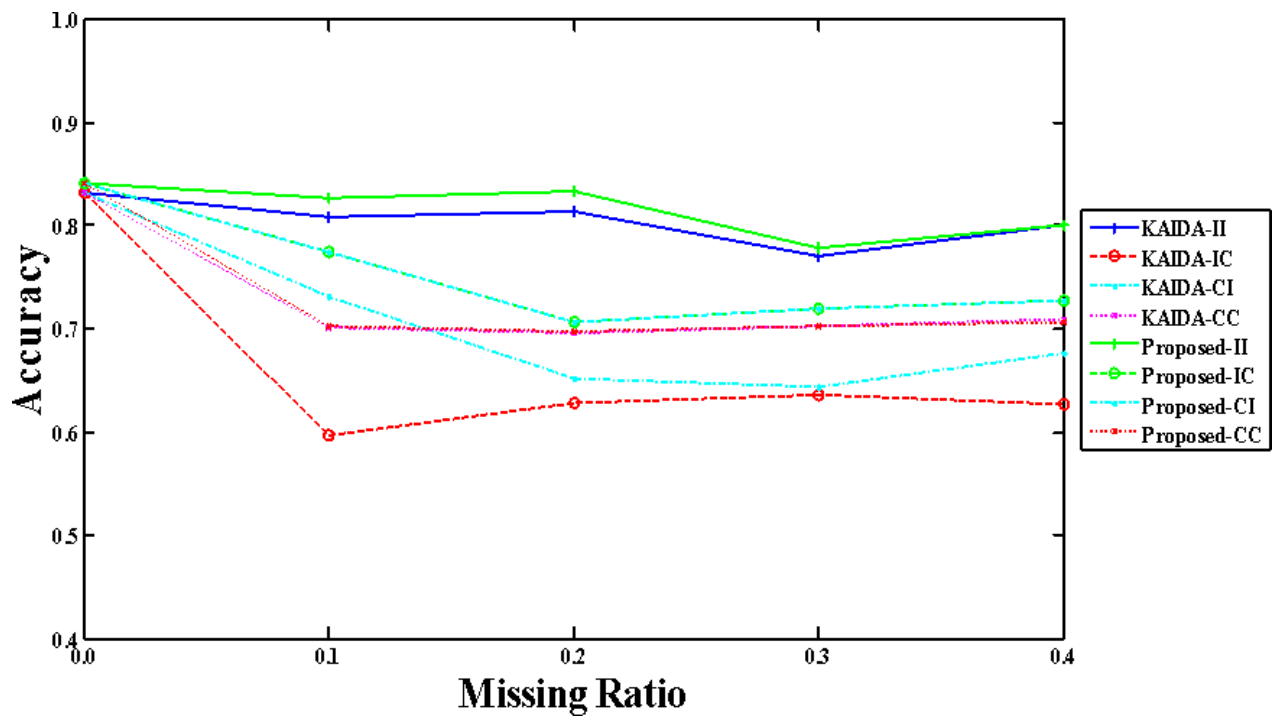

Fig. 2. ECG: Accuracy comparison between KAIDA and the proposed method with the linear kernel in intrinsic space. For the definitions of the abbreviations, please refer to the content in Section IV.

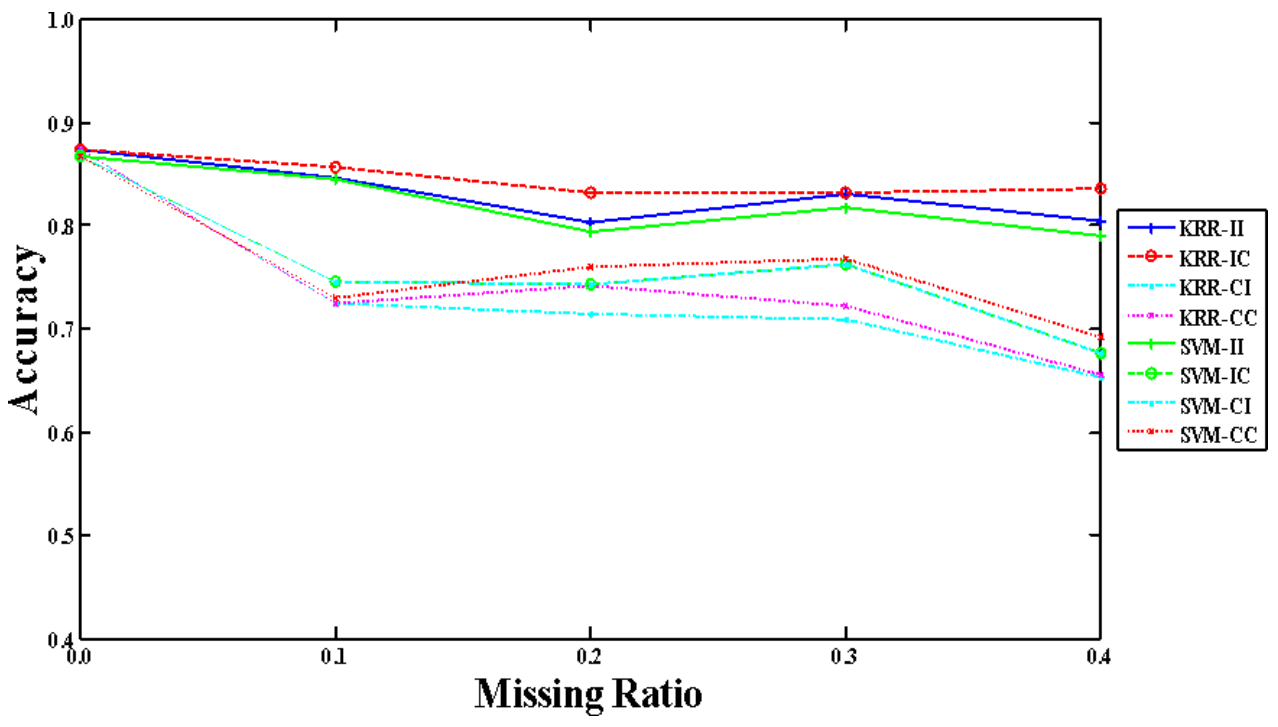

Fig. 3. ECG: Accuracy comparison between KRR and SVMs based on the proposed method with the linear kernel in intrinsic space. For the definitions of the abbreviations, please refer to the content in Section IV.

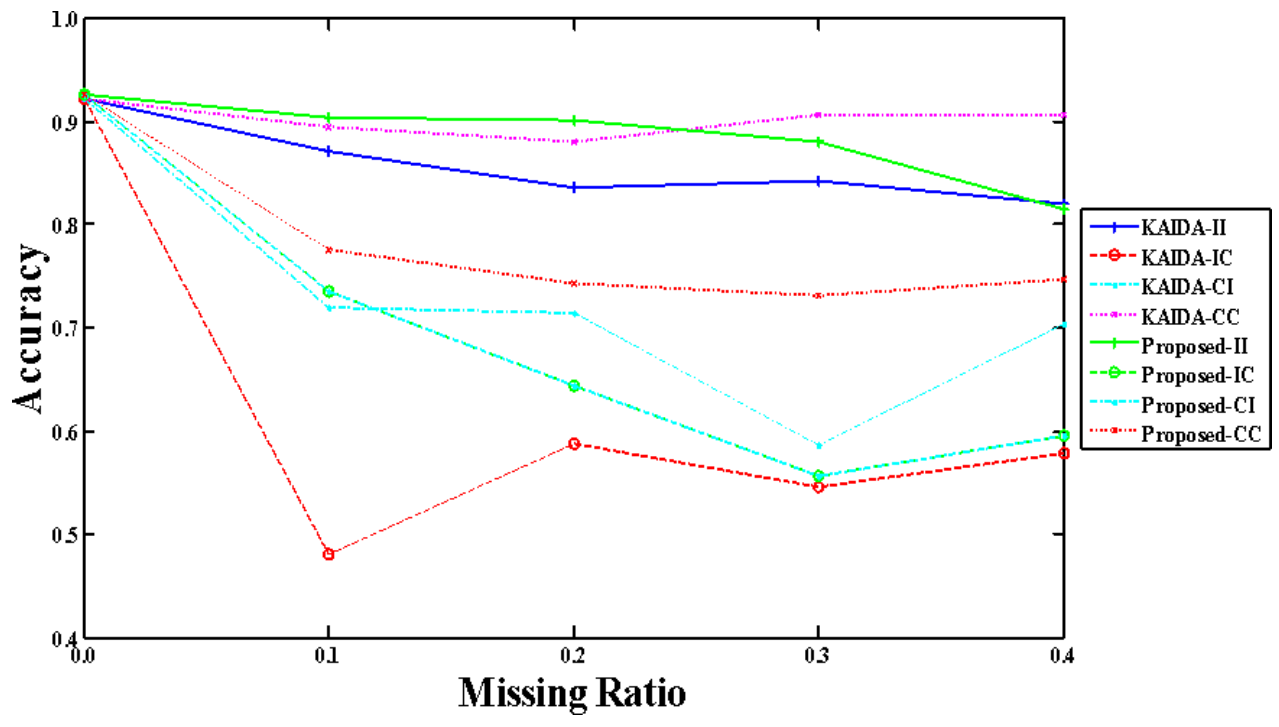

Fig. 4. ECG: Accuracy comparison between KAIDA and the proposed method with the poly3 kernel in intrinsic space. For the definitions of the abbreviations, please refer to the content in Section IV.

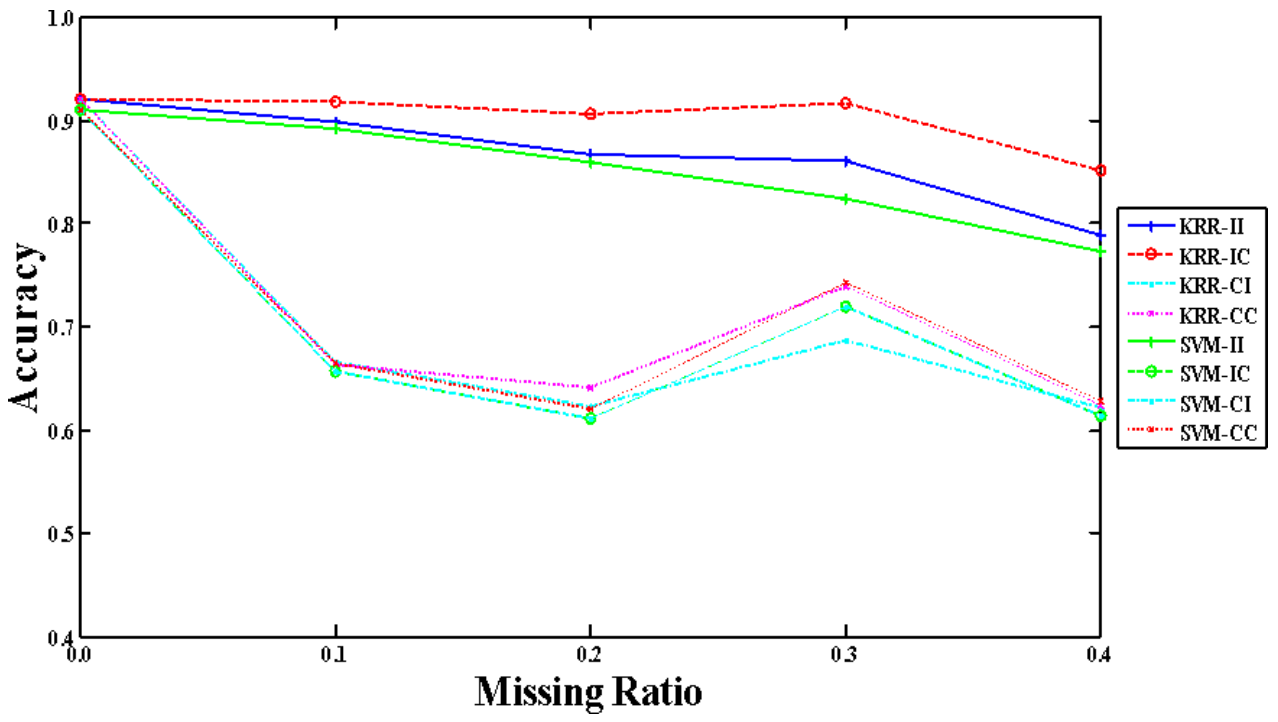

Fig. 5. ECG: Accuracy comparison between KRR and SVMs based on the proposed method with the poly3 kernel in intrinsic space. For the definitions of the abbreviations, please refer to the content in Section IV.

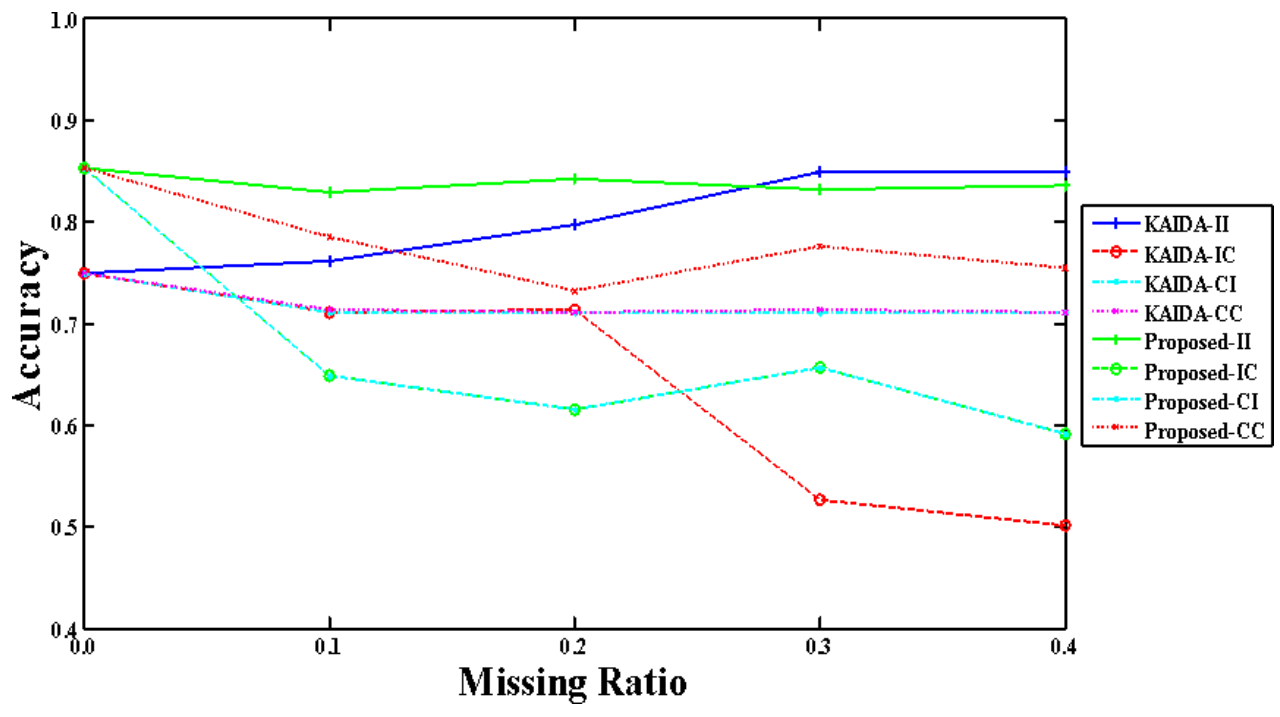

Fig. 6. ECG: Accuracy comparison between KAIDA and the proposed method with the RBF in empirical space. For the definitions of the abbreviations, please refer to the content in Section IV.

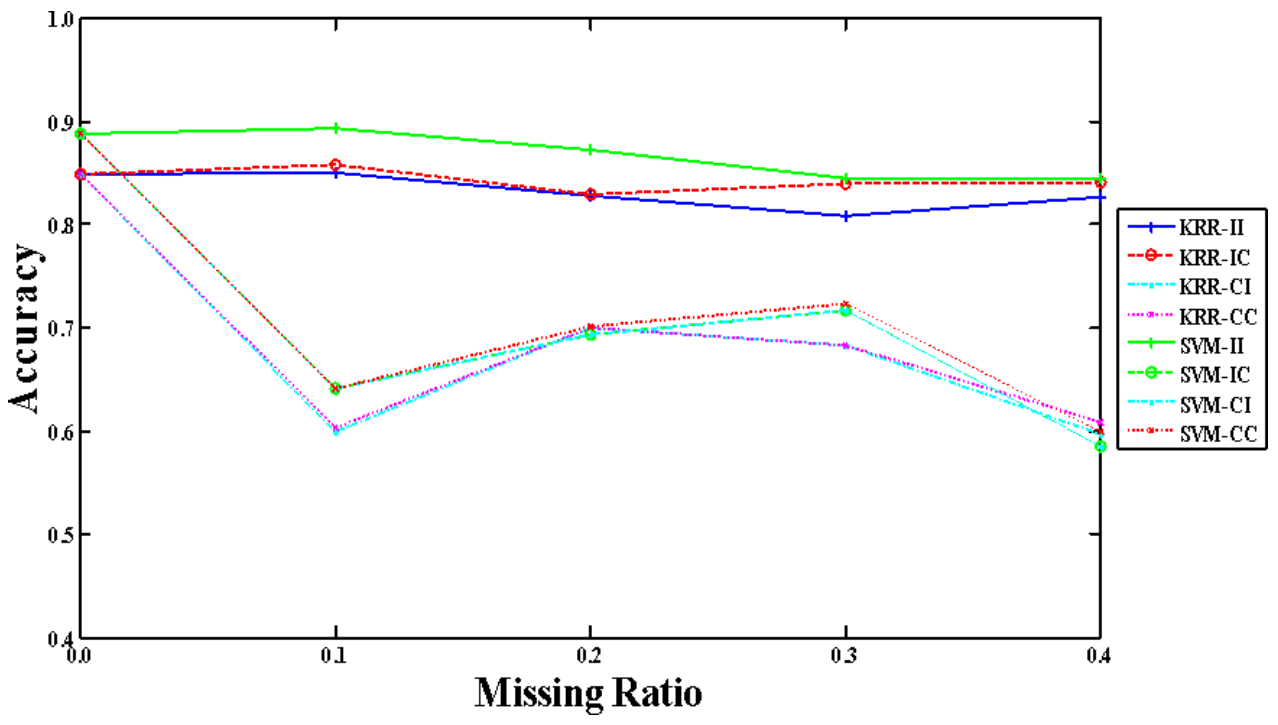

Fig. 7. ECG: Accuracy comparison between KRR and SVMs based on the proposed method with the RBF in empirical space. For the definitions of the abbreviations, please refer to the content in Section IV.

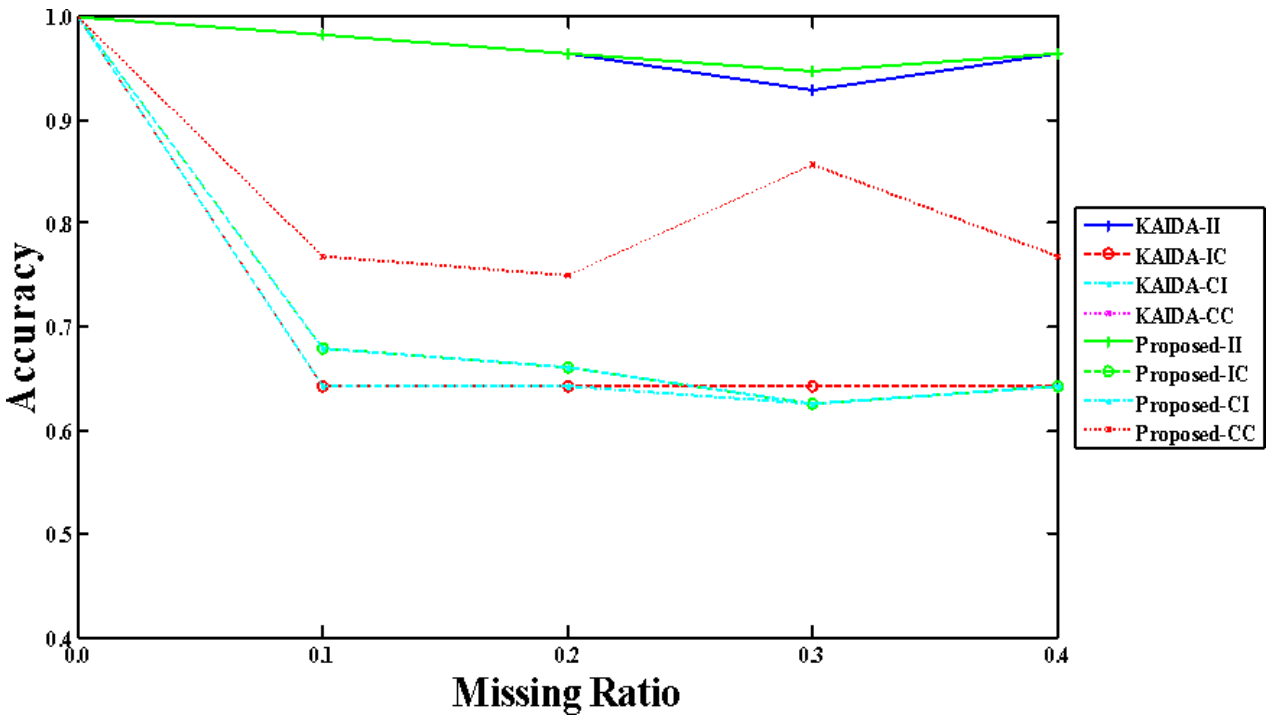

Fig. 8. ALL-AML: Accuracy comparison between KAIDA and the proposed method with the linear kernel in empirical space. For the definitions of the abbreviations, please refer to the content in Section IV.

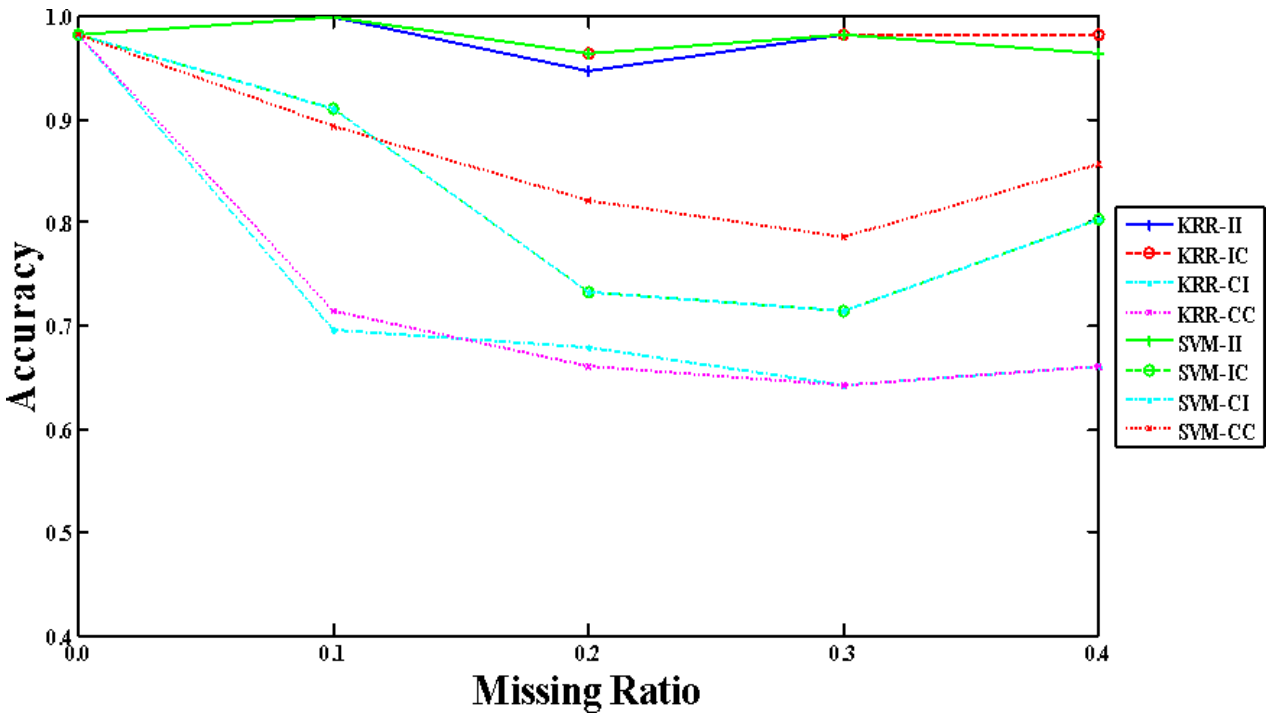

Fig. 9. ALL-AML: Accuracy comparison between KRR and SVMs based on the proposed method with the linear kernel in empirical space. For the definitions of the abbreviations, please refer to the content in Section IV.

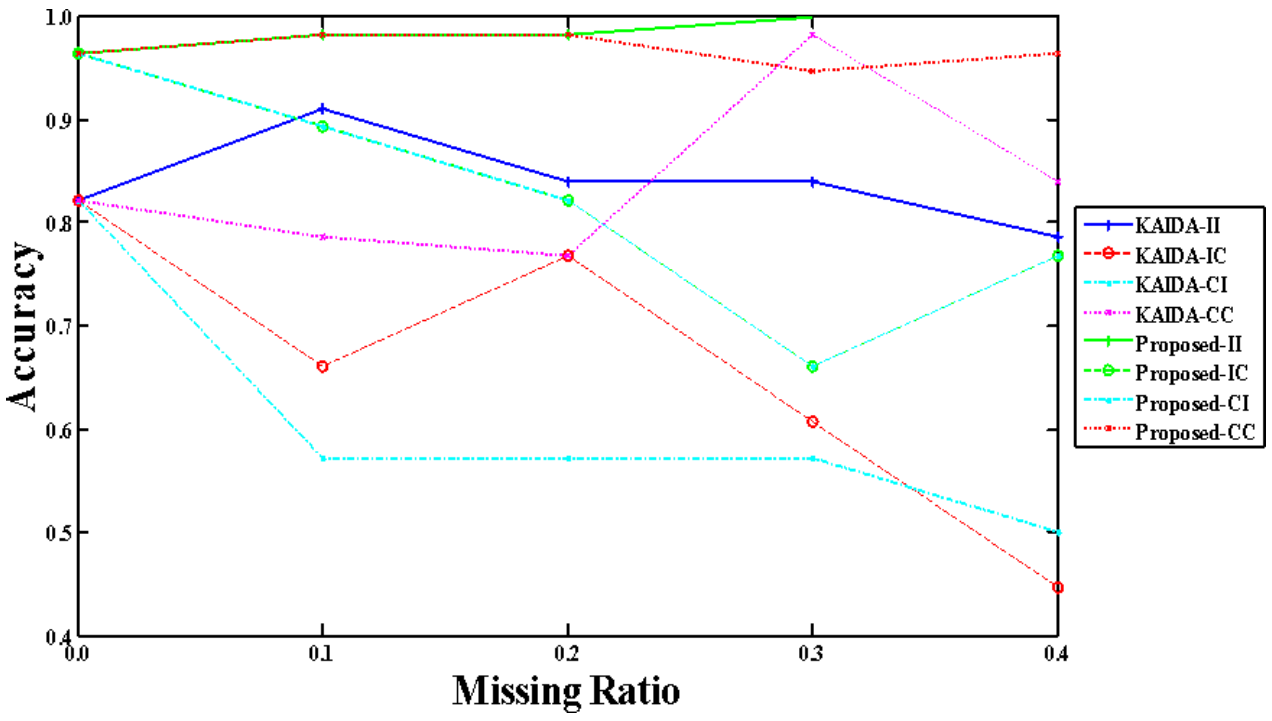

Fig. 10. ALL-AML: Accuracy comparison between KAIDA and the proposed method with the poly3 kernel in empirical space. For the definitions of the abbreviations, please refer to the content in Section IV.

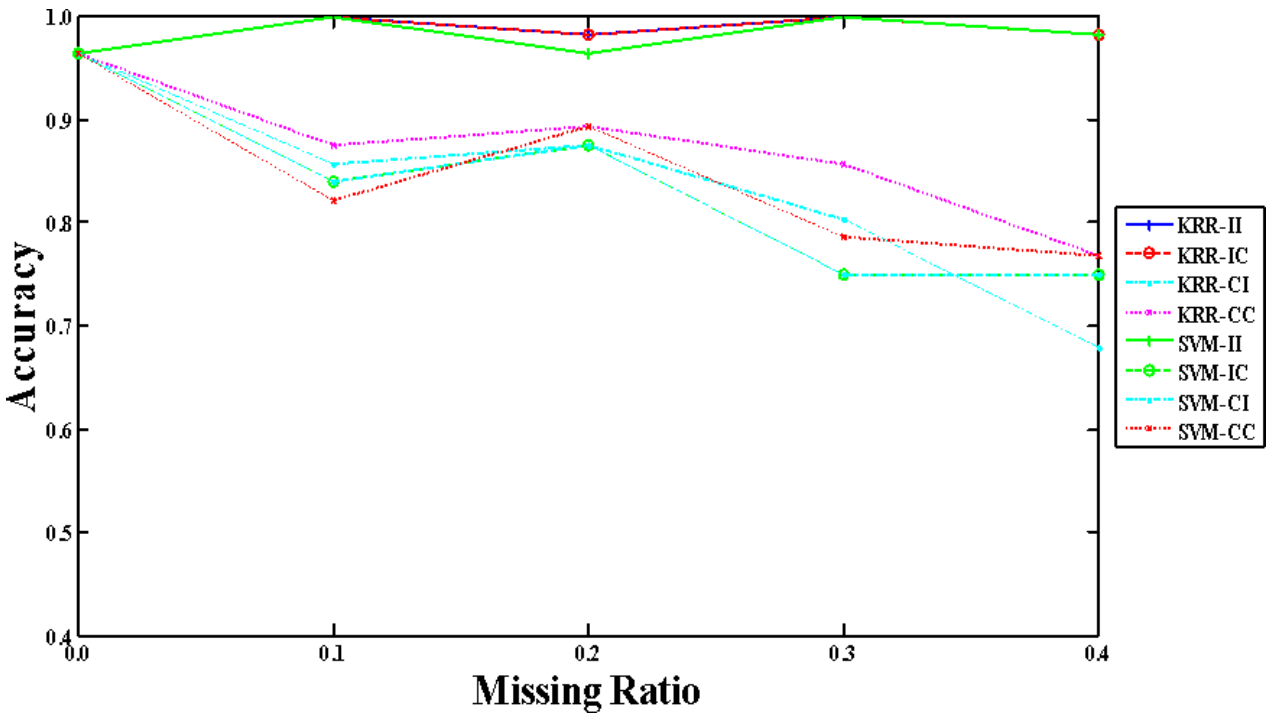

Fig. 11. ALL-AML: Accuracy comparison between KRR and SVMs based on the proposed method with the poly3 kernel in empirical space. For the definitions of the abbreviations, please refer to the content in Section IV.

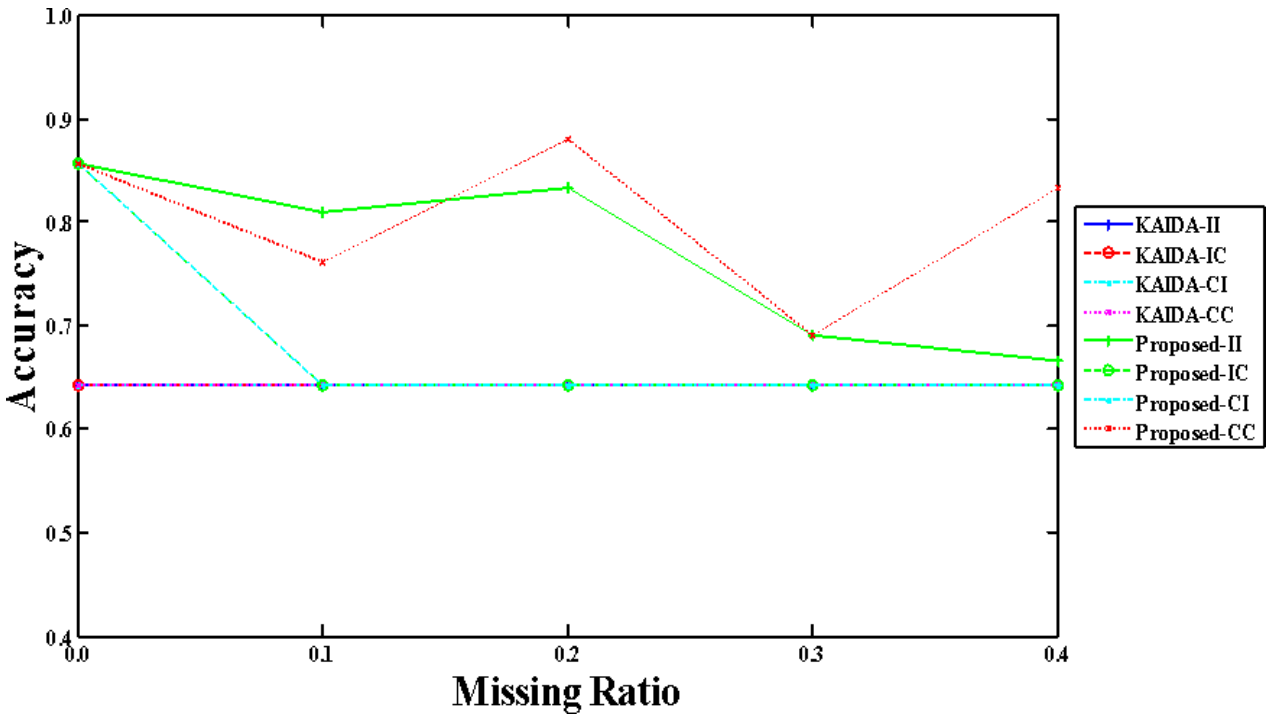

Fig. 12. ALL-AML: Accuracy comparison between KAIDA and the proposed method with the RBF in empirical space. For the definitions of the abbreviations, please refer to the content in Section IV.

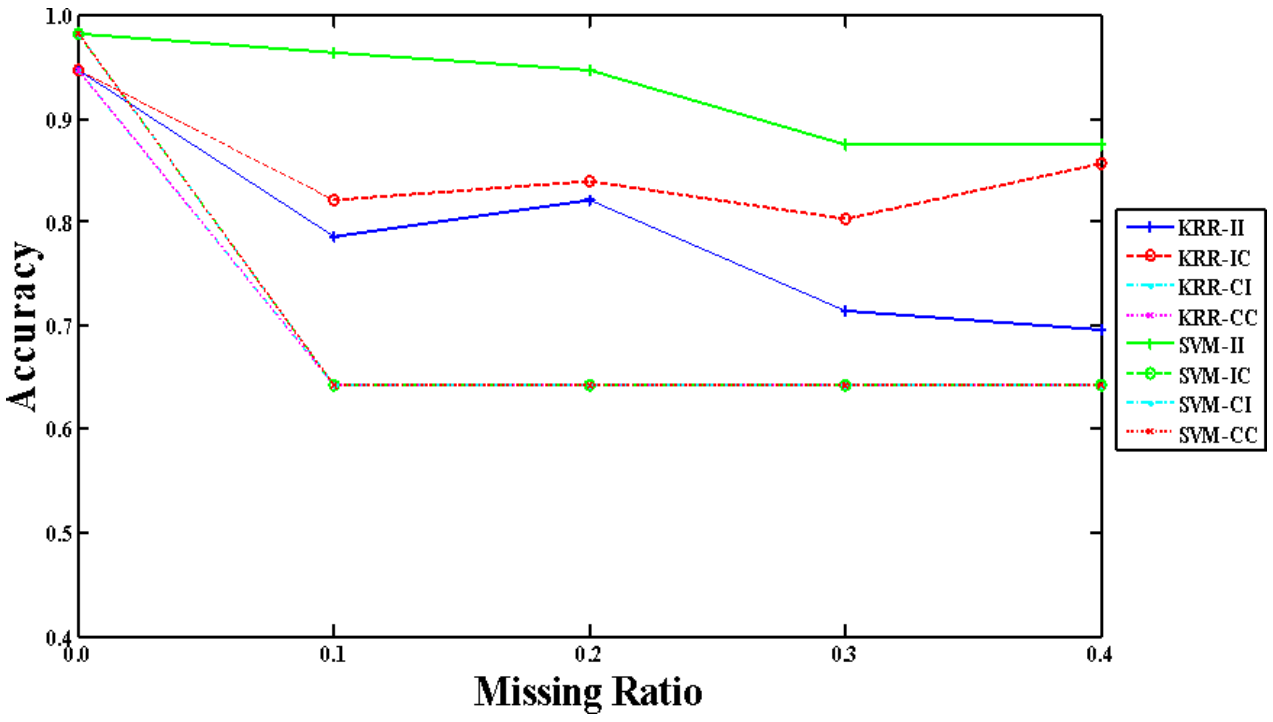

Fig. 13. ALL-AML: Accuracy comparison between KRR and SVMs based on the proposed method with the RBF in empirical space. For the definitions of the abbreviations, please refer to the content in Section IV.

Table I
Attributes of the Datasets

| Name | #Classes | #Instances | #Dimensions |
|---|---|---|---|
| ECG | 2 | 104033 | 21 |
| ALL-AML | 2 | 72 | 7129 |

Table II
Missing Values

| Name | Percentage of Missing Values | Mask Mode |
|---|---|---|
| Proposed | 00.00%–40.00% | Double Mask |
| KAIDA | 00.00%–40.00% | Double Mask |

Table III
Kernels

| Name | Kernels |
|---|---|
| Proposed | MPT Linear, MPT Poly3, & MPT RBFs |
| KAIDA | MPC, Masked Poly3, & Masked RBFs |

Table IV
Parameters of KRR and SVMs

| Name | Variance in Poly3 | Variance in RBFs | Ridge $\rho$ | Penalty $C$ |
|---|---|---|---|---|
| Proposed | 1.00 | 1.00 | 5.00 | 1.00 |
| KAIDA | 1.00 | 1.00 | 5.00 | 1.00 |

During the data imputation, 80.00% of the datasets were randomly selected for training, and the rest were used for testing. Two approaches — KAIDA [12] and our proposed method — were employed for assessment. The double mask was used in KAIDA and our method. The percentage of missing values ranged from 0.00% to 40.00%. Missing entries were randomly and uniformly generated. Notably, missing entries for training incomplete data and testing incomplete data were not exactly the same to guarantee randomness.

Regarding KRR, different kernels, including the linear kernel, the third-order polynomial (i.e., poly3) kernel, and RBFs, were applied to the MPC and MPT. The ridge parameter $\rho$ for KRR was 5.0. For high dimensional data, the top 200 dimensions with highest FDRs were selected during testing and training after imputation. Table II–Table IV respectively summarize the settings.

Fig. 2–Fig. 7 show the accuracy of KAIDA and our proposed method when ECG data were tested. Fig. 8–Fig. 13display the results of ALL-AML data. Each figure was independently generated as the training and testing subdatasets were randomly selected.

Symbols "I" and "C" respectively denote incomplete data and complete data (i.e., without missing values). When they are combined (i.e., II, IC, CI, and CC), the first symbol represents the training stage, and the last symbol signifies the testing phase. For example, "IC" means to perform training with complete data and to test the system with incomplete data.

Closely examining the results indicates that our proposed method generated higher accuracy, especially when high-order feature mapping functions were used, in contrast to KAIDA. However, both methods failed to maintain the accuracy when RBFs were used for testing ALL-AML data. The possible reason was that ALL-AML was high-dimensional data, and poly3 kernels already successfully delineated the distribution of the samples. The RBFs yielded overfitting hyperplanes. Table V–Table X summarize the numeric results of the figures. The accuracy in these tables was collected based on KAIDA and our proposed method, both with incomplete training data and incomplete testing data.

Another finding from our experimental results is the accuracy comparison between KRR and SVMs. When the linear and the poly3 kernels were used, the accuracy of KRR in classification was approximately near that of SVMs.

When RBFs were applied, SVMs were superior to KRR. Nevertheless, the computational time was more than that of KRR if $N > M$. Interestingly, we observed the accuracy of "II" was better than that of "CC" in some of the experimental results. The possible reason was overfitting caused by dimensionality, as earlier mentioned. Furthermore, we increased the discriminability of incomplete data by filling in with class-dependent averages. Based on the equation of Partial Fisher Discriminant-Ratios in (3), such a procedure enhanced classification accuracy.

Table V

Accuracy comparison using the ECG dataset and the linear kernel in intrinsic space

| **Missing Rate (%)** | **0** | **10** | **20** | **30** | **40** |
|---|---|---|---|---|---|
| **Proposed (%)** | 84.13 | 82.69 | 83.29 | 77.88 | 80.05 |
| **KAIDA (%)** | 83.17 | 80.89 | 81.37 | 77.04 | 80.05 |

Results with incomplete training data and incomplete testing data were selected

Table VI

Accuracy comparison using the ECG dataset and the poly3 kernel in intrinsic space

| **Missing Rate (%)** | **0** | **10** | **20** | **30** | **40** |
|---|---|---|---|---|---|
| **Proposed (%)** | 92.67 | 90.38 | 90.14 | 87.98 | 81.49 |
| **KAIDA (%)** | 92.19 | 87.14 | 83.65 | 84.25 | 82.09 |

Results with incomplete training data and incomplete testing data were selected

Table VII

Accuracy comparison using the ECG dataset and the RBF in empirical space

| **Missing Rate (%)** | **0** | **10** | **20** | **30** | **40** |
|---|---|---|---|---|---|
| **Proposed (%)** | 85.34 | 82.93 | 84.25 | 83.17 | 83.65 |
| **KAIDA (%)** | 75.00 | 76.20 | 79.81 | 84.98 | 84.98 |

Results with incomplete training data and incomplete testing data were selected

Table VIII

Accuracy comparison using the ALL-AML dataset and the linear kernel in empirical space

| **Missing Rate (%)** | **0** | **10** | **20** | **30** | **40** |
|---|---|---|---|---|---|
| **Proposed (%)** | 100.00 | 98.21 | 96.43 | 94.64 | 96.43 |
| **KAIDA (%)** | 100.00 | 98.21 | 96.43 | 92.80 | 96.43 |

Results with incomplete training data and incomplete testing data were selected

Table IX

Accuracy comparison using the ALL-AML dataset and the poly3 kernel in empirical space

| **Missing Rate (%)** | **0** | **10** | **20** | **30** | **40** |
|---|---|---|---|---|---|
| **Proposed (%)** | 96.43 | 98.21 | 98.21 | 100.00 | 100.00 |
| **KAIDA (%)** | 82.14 | 91.07 | 83.93 | 83.93 | 78.57 |

Results with incomplete training data and incomplete testing data were selected

Table X

Accuracy comparison using the ALL-AML dataset and the RBF in empirical space

| **Missing Rate (%)** | **0** | **10** | **20** | **30** | **40** |
|---|---|---|---|---|---|
| **Proposed (%)** | 85.71 | 80.95 | 83.33 | 69.05 | 66.67 |
| **KAIDA (%)** | 64.29 | 64.29 | 64.29 | 64.29 | 64.29 |

Results with incomplete training data and incomplete testing data were selected

## 5. Conclusion

This work presents a novel learning method for processing large-scale data with missing values. To effectively recover the discriminability of the data, this study proposes kernel-based data imputation, where the imputation focuses on discriminability enhancement of the multiside similarity between test and training samples. The proposed method takes advantage of teacher information in the training phase, where missing entries of the centroids are filled in with class-dependent substituted values. Then, a training sample is combined with its class centroid (generated without involving any clustering algorithm), and this correspondingly generates substituted values for training patterns. The equation shows that such imputation does not reduce Fisher Discriminant-Ratios, which indicate the discrepancy between two distributions. The proposed method can increase the discriminability of the training data by diminishing the variance. In terms of SNRs, the noise is minimized. Although a test sample still uses masks to zero-pad its missing entries, the three-side kernel matrix compensates the drawback of zero-padding.

Experiments were conducted by evaluating two open datasets. Each dataset represents one type of big data. The result showed that the difference in accuracy between the proposed method and the baseline increased when high-order kernels were used. Besides, our proposed method yielded better results than the baseline did. Such findings indicated that the proposed methods were more effective than the baseline for handling missing data, thereby demonstrating the feasibility of the proposed idea.

## Acknowledgements

Dr. Chen would like to thank Prof. Sun-Yuan Kung for his supervision and teaching when Dr. Chen worked as a postdoctoral fellow at Princeton University, USA.